\documentclass[letterpaper, 10 pt, conference]{ieeeconf}  % Comment this line out if you need a4paper

\IEEEoverridecommandlockouts                              % This command is only needed if
\usepackage{graphicx}
\usepackage{amssymb}
\usepackage{amsmath}
\usepackage{bm}
\usepackage{xspace}
\usepackage{makecell}
\usepackage{booktabs}
\usepackage{url}
\usepackage{subfigure}
\usepackage{tabularx}
\usepackage{multirow}
\usepackage{multicol}
\usepackage{threeparttable}
\usepackage{placeins}
\usepackage{balance}
\usepackage{marvosym}

\usepackage{cite}
\usepackage{caption}
\makeatletter
\let\NAT@parse\undefined
\makeatother
\usepackage[colorlinks,urlcolor=blue]{hyperref}

\usepackage{color}
\usepackage{xcolor}

\title{\LARGE \bf
Dissecting Advantage-Guided Post-Training for \\ Vision-Language-Action Policies
}
\author{Jiahang Cao$^{1}$$^\dagger$$^\star$, Hanye Zhao$^{1}$$^\dagger$$^\star$, Hang Lai$^{2}$$^\text{\Letter}$, Shenyu Zhang$^{2}$, Xiaoshen Han$^{1}$$^\dagger$, Xinghang Li$^{2}$, \\
Futeng Liu$^{2}$, Wanli Peng$^{2}$, Heyun Wang$^{2}$, Yunhong Wang$^{2}$, Jason Li$^{2}$, Yong Yu$^{1}$, Weinan Zhang$^{1}$$^\text{\Letter}$
\thanks{$^\star$Equal contribution.\quad $^\text{\Letter}$Corresponding author.}
\thanks{$^\dagger$Work done during internship at Xiaomi Robotics.}
\thanks{$^{1}$School of Computer Science, Shanghai Jiao Tong University.}
\thanks{$^{2}$Xiaomi Robotics.}
}

\begin{document}
\bstctlcite{IEEEexample:BSTcontrol}

\maketitle

\thispagestyle{empty}
\pagestyle{empty}

\begin{abstract}

Advantage-guided reinforcement learning provides a practical way to post-train vision-language-action (VLA) policies using limited robot data. 
However, its performance depends on several coupled choices, including how critic-derived advantages are constructed, calibrated, and used for policy training. 
Existing recipes often combine these choices into a single end-to-end procedure, making their individual effects difficult to identify. 
In this work, we dissect advantage-guided VLA post-training through a controlled empirical study that separates these design choices while accounting for their distinct estimands. 
We develop stage-specific offline evaluation methods to screen alternative choices efficiently, without requiring extensive real-robot policy evaluations for every possible combination. 
The staged evaluation identifies a modular recipe that combines temporal-difference advantage construction, group-wise calibration, and continuous advantage weighting.
Across four real-world bimanual tasks, the resulting recipe improves mean task progress and success over the SFT initialization by 0.42 and 0.63, respectively.
% The selected modular recipe achieves higher success rates than commonly used baselines on four real-world manipulation tasks.
Moreover, the proposed evaluation diagnostics show an overall alignment with downstream real-world performance, supporting their use for interpreting empirical outcomes and selecting advantage-guided post-training designs in practice.
% Experiments on hold-out simulator tasks provide complementary evidence that several relative effects recur beyond the real-world setting.
Videos are available at our project website \href{dissectvla.github.io}{dissectvla.github.io}.

\end{abstract}

% Contents
\section{Introduction}
\label{sec:introduction}

Vision-language-action (VLA) models provide a general interface for language-conditioned robot control~\cite{rt2, openvla, pi0, black2025pi05}, but effectively refining these policies from limited and heterogeneous robot data remains challenging, particularly when feedback is sparse over long-horizon tasks. 
A post-training dataset may combine demonstrations, autonomous rollouts, and trajectory segments collected around human interventions, all generated by different behavior policies~\cite{zhou2024soar,jiang2024transic,xu2025rldg}. 
Such data contain successful behavior, uncorrected failures, and corrective or recovery actions, making it important to determine how different data samples should contribute to policy learning. 
Equal weighting may expose the actor to failures as if they were demonstrations, whereas overly aggressive filtering may discard corrective or recovery behavior. 
These considerations motivate methods that can modulate the contribution of individual samples during policy learning.

\begin{figure}[t]
    \centering
    \includegraphics[width=0.98\linewidth]{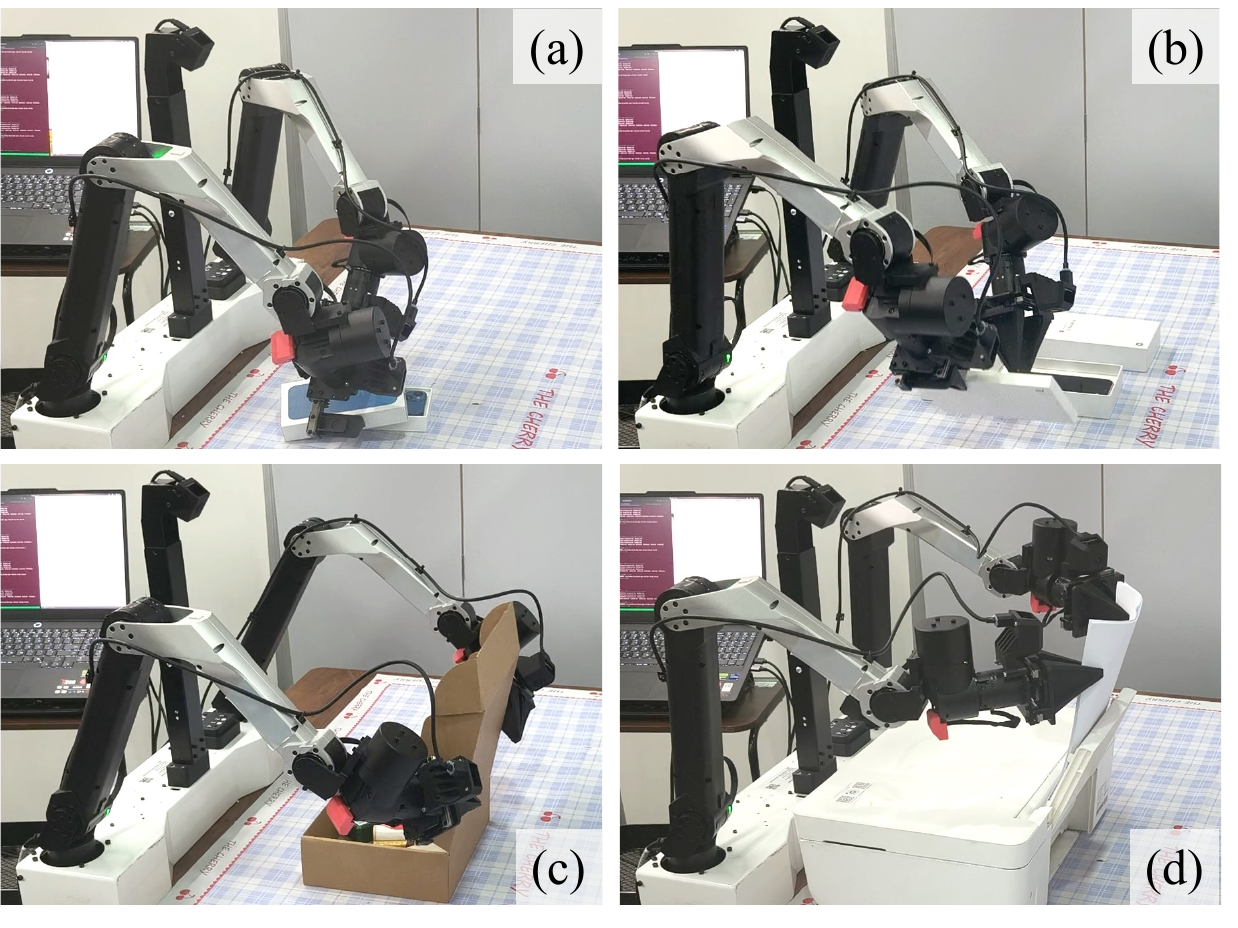}
    \caption{Representative scenes from the four
    real-world manipulation tasks in our evaluation.
    (a) Pack Phone Easy, 
    (b) Pack Phone Hard, 
    (c) Pack Box, 
    and (d) Refill Printer Paper. 
    % Each panel shows a representative scene from the corresponding task.
}
    \label{fig:realscene}
    \vspace{-15pt}
\end{figure}

As one such approach, advantage-guided reinforcement learning (RL) has attracted growing interest and has been increasingly applied to VLA post-training ~\cite{tan2025riptvla,chen2025tgrpo,yang2026aloe,shu2025rftf,ma2025divo}. 
By using critic-derived advantage to reweight or filter sampled action chunks, it can exploit mixed-quality experience without requiring the dataset to consist entirely of expert demonstrations or relying on extensive additional online interaction. 
Typically, an advantage-guided post-training pipeline involves three stages: advantage construction, advantage calibration, and advantage utilization. 
These stages address different questions: construction determines what quantity the advantage estimates, calibration determines how the advantage is made comparable across samples, and utilization determines how it shapes policy updates through weighting or filtering. 
Each stage also admits multiple design choices, yielding a large space of possible post-training recipes.
% These stages jointly determine how critic-derived signals influence policy learning, and therefore their effects cannot be reliably inferred from existing end-to-end recipes alone.

Because choices across stages jointly shape how advantage influences policy learning, performance differences between existing end-to-end recipes cannot be reliably attributed to any individual stage. 
It remains unclear which choices within each stage are most effective, and why these choices are responsible for observed policy improvements.
Changing one stage may also alter the sample weights, sample exposure, and effective sample size produced by the complete pipeline. 
% Exhaustively evaluating all possible combinations would further require substantial policy training and real-robot evaluation, making direct comparison costly in data-limited settings. 
Comparing many such combinations would require substantial policy training and evaluation, which is costly in real-robot settings~\cite{snyder2025policycomparison,anwar2025activeeval}.

\begin{figure*}[htb]
    \centering
    \includegraphics[width=\textwidth]{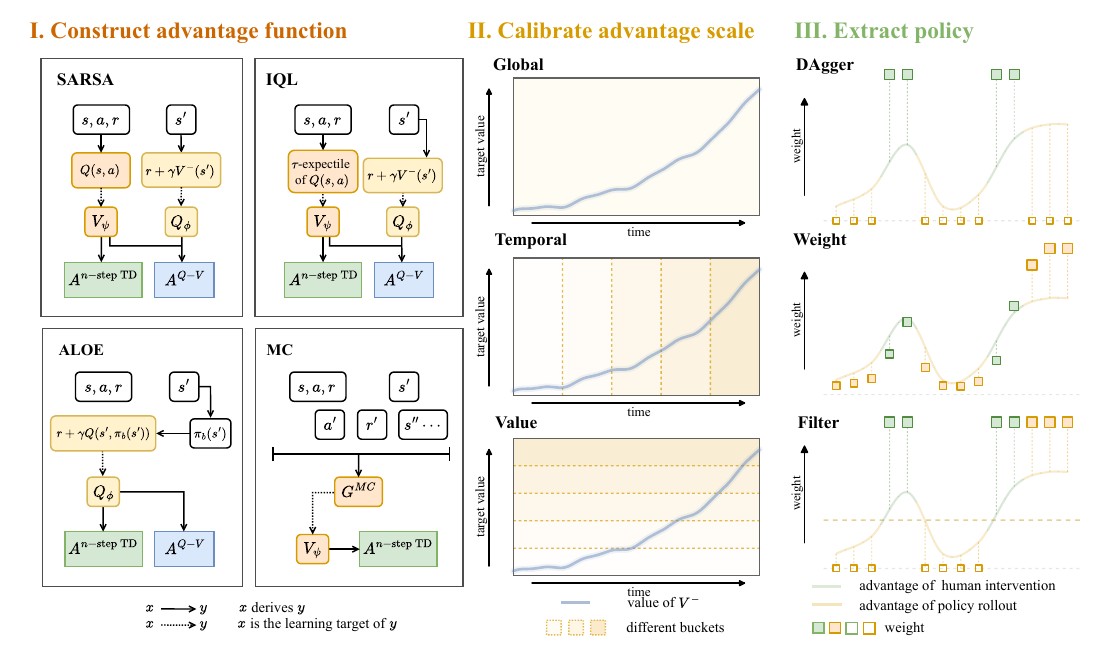}
    \vspace{-20pt}
    \caption{
Overview of our three-stage evaluation of advantage-guided VLA post-training. 
Stage I compares different advantage construction schemes,
Stage II varies the grouping rule in advantage calibration, 
and Stage III examines three sample-weighting strategies in real-robot policy training.
}
\label{fig:teaser}
\vspace{-18pt}
\end{figure*}

To this end, we dissect the post-training pipeline through a sequential three-stage evaluation protocol that screens design choices stage by stage before assessing their combined effect at the policy level. As shown in Fig.~\ref{fig:teaser},
stage I compares critic-training and advantage-estimation choices using offline sample-ordering diagnostics. 
Stage II fixes the selected Stage-I configuration and compares calibration choices using offline distributional diagnostics. 
Stage III then fixes the selected construction and calibration configurations and evaluates their use in policy training using the \(\pi_{0.5}\) flow-matching actor as a representative VLA policy backbone~\cite{black2025pi05}. 
The first two stages provide offline screening evidence, whereas the third provides the policy-level evidence by comparing complete policy-training conditions. 
By propagating the selected configuration from one stage to the next, this protocol avoids retraining and evaluating every possible cross-stage combination on real robots. 
We apply the protocol to offline real-robot datasets and evaluate the resulting policies on four real-world manipulation tasks that span object insertion, long-horizon sequencing, multi-object interaction, and deformable-object handling (Fig.~\ref{fig:realscene}).
This real-robot study complements prior empirical studies centered on simulation and cost-free settings~\cite{rft-survey,liu2025rlvla,zang2026rlinfvla,dong2026batchonlinerl}.

Across the three stages, the results identify a consistent pattern for selecting and using advantages. 
In Stage I, $n$-step TD constructions outperform chance on both sample-ordering diagnostics, whereas $Q - V$ and Monte Carlo constructions are generally weaker. 
In Stage II, value-based calibration produces the largest average reduction in cross-sub-task advantage imbalance. 
With the selected construction and calibration, continuous advantage weighting performs best on both task progress and final task success. 
Real-world ablation experiments on alternative construction and calibration choices show an overall alignment between downstream policy performance and the corresponding offline diagnostics, supporting the use of these diagnostics to interpret and select post-training designs without exhaustively evaluating every possible combination. 

In summary, this study makes three contributions:
\begin{itemize}
    \item We develop a sequential evaluation protocol with stage-specific offline diagnostics to screen post-training choices and reduce costly real-robot evaluations.
    \item Guided by this protocol, we derive a simple modular post-training recipe that combines effective choices for advantage construction, calibration, and utilization.
    % \item We demonstrate the recipe’s effectiveness on both real-world and simulated tasks, and establish a strong correlation between offline diagnostics and real-world policy performance.
    \item We demonstrate the recipe’s effectiveness on real-world tasks, and illustrate a consistent relationship between offline diagnostics and real-world policy performance.
\end{itemize}

\section{Related Work}
\label{sec:related_work}
\noindent\textbf{VLA Post-Training.}
Adapting a pretrained VLA policy to task-relevant robot experience generally follows two complementary paradigms: supervised fine-tuning (SFT) and reinforcement fine-tuning (RFT).
% \paragraph{Reinforcement Learning for VLA Policies}
SFT methods optimize a supervised objective that directly matches the policy to demonstrated or corrected actions~\cite{openvla,openvla-oft}, whereas RFT methods optimize the policy toward higher-reward behavior using reward or value feedback~\cite{chen2025conrft,tan2025riptvla, shu2025rftf, rft-survey,hu2025flare,guo2025irevla}. %,lv2026fpo}. 
Empirical comparisons further report improved semantic and execution robustness over SFT~\cite{liu2025rlvla}.
In real-robot settings, post-training data are often limited, and autonomous rollouts may contain failures from which the policy cannot recover without intervention~\cite{hil-serl, zhao2025more,huang2026nonexpert}. 
Human takeover data are therefore commonly used as corrective supervision, providing recovery trajectories and examples from task-relevant states. 
As an interactive SFT method, HG-DAgger~\cite{kelly2019hgdagger} extends the dataset-aggregation principle of DAgger~\cite{ross2011dagger} to interactive human interventions during autonomous execution. 
HG-DAgger can provide corrective behavior for states encountered during execution, but its original formulation assigns uniform supervision to collected human-intervention data and does not directly exploit correct actions produced during autonomous execution. 
Our study also incorporates human-takeover data, but focuses on RFT, where critic-derived advantages can differentiate the training influence of heterogeneous dataset samples~\cite{yang2026aloe,huang2025corft}.

\noindent\textbf{Advantage-Guided Reinforcement Learning.}
Value or advantage estimates can provide non-uniform supervision for policy improvement~\cite{mao2026arm, su2026igrft, dong2026batchonlinerl}.
Early offline RL methods mainly differ in how they estimate advantages and translate them into policy updates: 
IQL fits the state-value function by expectile regression on in-sample action values and performs exponentially advantage-weighted behavioral cloning using $Q-V$~\cite{kostrikov2022iql}, while AWR fits a state-value baseline to empirical Monte Carlo returns and applies exponential advantage weighting~\cite{peng2020awr}. 
Building on this principle, CRR supports both binary and continuous critic-guided regression~\cite{wang2020crr}, and AWAC extends advantage-weighted updates from offline pretraining to online refinement~\cite{nair2020awac}. 
Recent VLA studies adapt these ideas to high-dimensional, temporally extended control by constructing relative trajectory advantages~\cite{tan2025riptvla,chen2025tgrpo} or aligning temporal-difference learning with action chunking~\cite{huang2025corft}. 
In particular, ALOE learns a $Q$-function with policy-sampled chunked TD targets for action-level off-policy evaluation, then derives $Q-V$ advantages for weighted policy improvement~\cite{yang2026aloe}.
RECAP instead learns a distributional value function fitted to Monte Carlo returns and conditions the policy on a binarized advantage indicator~\cite{pi06-star}. 
% For long-horizon manipulation, ARM further learns relative advantage rewards from progress labels to reweight offline actions~\cite{mao2026arm}, whereas IG-RFT combines interaction-guided AWR with hybrid rewards and human-in-the-loop training~\cite{su2026igrft}. 
Despite their effectiveness, these methods couple advantage construction and utilization under different data and interaction settings, making their individual effects difficult to isolate. 
Our work addresses this gap by separating construction, calibration, and utilization under a shared actor, dataset, and training budget.

% \paragraph{Advantage-guided Reinforcement Learning}
% Advantage estimates provide a common interface for using value information to guide policy improvement across offline and interactive reinforcement learning.
% In offline RL, IQL estimates a state-value function through expectile regression and extracts a policy by advantage-weighted behavioral cloning~\cite{kostrikov2022iql}.
% AWR maps estimated advantages to exponential weights for supervised policy learning~\cite{peng2020awr}, AWAC uses advantage-weighted actor updates to combine offline data with subsequent online interaction~\cite{nair2020awac}, and CRR uses critic-derived action-quality estimates to construct binary or exponential regression weights~\cite{wang2020crr}.
% Advantage-based policy optimization is also used in VLA post-training, including leave-one-out advantage estimation with sparse binary rewards~\cite{tan2025riptvla}, fused trajectory- and step-level relative advantages~\cite{chen2025tgrpo}, and action-level off-policy evaluation followed by advantage-weighted policy improvement~\cite{yang2026aloe}.
% Our empirical study brings these related design choices into a common VLA post-training pipeline and evaluates advantage construction, calibration, and utilization as separate stages.

\section{Preliminaries}
\label{sec:preliminaries}
VLA policy post-training can be formulated as a Markov Decision Process (MDP), denoted by $\mathcal{M}=(\mathcal{S},\mathcal{A},\mathcal{R},\mathcal{P},\gamma)$, where $\mathcal{S}$ is the state space, $\mathcal{A}$ is the action space, $\mathcal{R}$ is the reward function, $\mathcal{P}$ denotes the transition dynamics, and $\gamma$ is a discount factor.
A transition sampled from the dataset is represented as $T=(s_t,\mathbf{a}_t,r_t,s_{t+1},\mathbf{a}_{t+1})$. 
$s_t$ denotes the policy input at decision step $t$, including the natural language instruction, visual observations, and robot state.
$\mathbf{a}_t=(a_{t,0},a_{t,1},\ldots,a_{t,C-1})$ denotes the action chunk, where $a_{t,i}\in\mathcal{A}$ and $C$ is the chunk length. 
The $C$ primitive actions are executed consecutively before the next policy query, reducing the decision frequency.
Consequently, $s_{t+1}$ denotes the state reached $C$ primitive timesteps after $s_t$. 
To avoid task-specific dense reward engineering and facilitate scalable post-training, we adopt a sparse reward such that $r_t=1$ only for the terminal transition of a successful episode, and $r_t=0$ otherwise. An indicator $d_t\in\{0,1\}$ denotes whether the transition terminates an episode.

\subsection{Advantage Construction}
Advantage construction consists of critic learning followed by advantage estimation. For TD-based critics, the action-value function $Q_\phi$ is learned using the Bellman target:
\begin{equation}
\begin{aligned}
y_t
&=
r_t+\gamma(1-d_t)V_{\psi^-}(s_{t+1}),\\
\mathcal{L}_{Q}(\phi)
&=
\mathbb{E}_{T\sim\mathcal{D}}
\left[
\operatorname{MSE}\!\left(
Q_{\phi}(s_t,\mathbf{a}_t),
\operatorname{sg}[y_t]
\right)
\right],
\end{aligned}
\label{eq:value_loss}
\end{equation}
where $\operatorname{sg}$ stops gradients and $V_{\psi^-}$ denotes the target value network, whose parameters are updated as an exponential moving average of those of $V_\psi$. The learning of $V_\psi$ is method-dependent: it may be fitted by expectile regression over $Q$-values~\cite{kostrikov2022iql} or, in the MC variant, directly to empirical returns $G_t^{\mathrm{MC}}$~\cite{pi06-star}. Alternatively, the continuation value may be obtained from $Q$ at a policy-sampled action~\cite{yang2026aloe}.

% The losses for learning the action-value network $Q_\phi$ and the state-value function $V$ have a unified form as follows. 
% Let $F\in\{Q,V\}$, with inputs $x_t^Q=(s_t,\mathbf{a}_t)$ and $x_t^V=s_t$. 
% Let $\mathcal{T}_F$ denote the Bellman operator associated with the selected continuation-value backup.
% The Bellman target and the corresponding value-function loss are
% \begin{equation}
% \begin{aligned}
% y_t^F
% &=
% \bigl(\mathcal{T}_F F_{{\theta}^{-}}\bigr)(x_t^F),\\
% \mathcal{L}_\mathrm{value}(\theta)
% &=
% \mathbb{E}_{T\sim\mathcal{D}}
% \left[
% \rho_F\!\left(
% F_{\theta}(x_t^F)
% -
% \operatorname{sg}\!\left[y_t^F\right]
% \right)
% \right]
% \end{aligned}
% \label{eq:value_loss}
% \end{equation}
% where $F_\theta\in\{Q_\phi,V_\psi\}$ and $\theta^-$ denotes the parameters used to construct the target,
% $\operatorname{sg}[\cdot]$ denotes stop-gradient, 
% and $\rho_F$ is the regression loss.

%% advantage calculation

% The $\pi_{0.5}$ actor assigns replay example $i$ the conditional flow-matching loss $\ell_{\mathrm{FM},i}(\theta)$~\cite{black2025pi05,lipman2023flowmatching}

From the learned values, we consider two advantage estimates used in offline RFT~\cite{yang2026aloe,huang2025corft}. The $Q-V$ estimate compares the logged action with its state-value baseline:
\begin{equation}
A_t^{Q-V}
=
Q_{\phi}(s_t,\mathbf{a}_t)
-
V_{\psi}(s_t).
\label{eq:advantage_qv}
\end{equation}

The $n$-step TD estimate compares an $n$-step bootstrapped return with the current state value:
\begin{equation}
\resizebox{0.9\linewidth}{!}{%
$
\begin{gathered}
A_t^{\mathrm{n\text{-}step\ TD}} = R_t^n + \gamma^n \mathsf{m}_{t,n}V_{\psi^-}(s_{t+n}) - V_{\psi}(s_t), \\
R_t^n = \sum_{j=0}^{n-1} \gamma^j \mathsf{m}_{t,j} r_{t+j}, \quad \mathsf{m}_{t,j} = \prod_{k=0}^{j-1}(1-d_{t+k}), \ \mathsf{m}_{t,0}=1.
\end{gathered}
$
}
\label{eq:advantage_n_step_td}
\end{equation}
Since $t$ indexes action chunks, its rewards, discounting, termination masks, and bootstrap state share the same chunk-level time scale~\cite{li2025actionchunking,huang2025corft}.

\subsection{Advantage Calibration}
Advantages are commonly normalized within a batch or dataset to stabilize policy optimization~\cite{schulman2017proximal}. Recent RL post-training methods further normalize advantages within structurally homogeneous groups for finer-grained control~\cite{zhu2025stratifiedgrpo}. We present group-wise normalization as the general form, with standard normalization recovered by assigning all samples to a single group. Let $\mathcal{G}$ denote a grouping rule and let $g_i=\mathcal{G}(i)$ be the group assigned to example $i$.
The corresponding calibration function is
\begin{equation}
\begin{aligned}
z_i &= \frac{u_i-\mu_{g_i}}{\max\!\left(\sigma_{g_i},\sigma_{\min}\right)},\\
\mu_g = \mathbb{E}_{j:g_j=g}[u_j]&, \quad \sigma_g^2 = \mathbb{E}_{j:g_j=g} \left[(u_j-\mu_g)^2\right],
\end{aligned}
\label{eq:score_calibration}
\end{equation}
where $u_i$ and $z_i$ denote the raw and calibrated advantages, respectively.
% The grouping rule $\mathcal{G}$ determines the calibration operator by specifying how examples are assigned to groups.
A scale floor $\sigma_{\min}$ prevents numerical instability when the within-group variance is small.
% These calibration groups should be distinguished from using value estimates to evaluate individual actions.  

\subsection{Advantage Utilization}

After calibration, each training example $i$ is assigned a calibrated advantage $z_i$.
Let $\ell_i(\theta)$ denote the supervised policy loss for example $i$, which is the flow-matching loss used by the $\pi_{0.5}$ actor in our study~\cite{black2025pi05,lipman2023flowmatching}.
Given a nonnegative training weight $w_i$, the weighted actor objective for a training batch $\mathcal{B}$ is
\begin{equation}
\mathcal{L}_{\pi}(\theta) = \frac{\sum_{i\in\mathcal{B}}w_i\ell_i(\theta)}{\sum_{i\in\mathcal{B}}w_i},
\qquad \sum_{i\in\mathcal{B}}w_i>0.
\label{eq:weighted_actor_loss}
\end{equation}
The weight $w_i$ is determined by the calibrated advantage $z_i$ and the selected utilization rule.
\section{Study Design}
\label{sec:method}

We decompose advantage-guided policy post-training into three sequential stages---advantage construction, calibration, and utilization---to isolate the role of each component under a fixed offline dataset and actor setup. 
Exhaustively training and evaluating all cross-stage combinations on real robots is prohibitively expensive; we therefore screen construction and calibration with offline diagnostics in Stages I and II, and reserve actor training and deployment for Stage III. 
This ordering follows the pipeline dependency: calibration operates on constructed advantages, and utilization consumes calibrated advantages. 
We test whether the offline preferences are reflected in downstream policy performance by revisiting alternative construction and calibration choices in real-robot ablations. 
Fig.~\ref{fig:teaser} summarizes this staged design and the variants considered in each stage. 
Accordingly, we study the following research questions: 
\textbf{RQ1 (advantage construction):} Which combinations of critic-learning scheme and advantage construction produce advantages that reliably recover the expected quality orderings among dataset actions?
\textbf{RQ2 (advantage calibration):} How does advantage calibration affect the comparability of advantages across sub-tasks in the fixed dataset?
\textbf{RQ3 (advantage utilization):} How do different strategies for utilizing advantages affect policy performance across real-robot tasks?

\subsection{Stage I: Advantage Construction}
Stage I compares four representative value-learning schemes and evaluates seven advantage construction configurations.
The value-learning schemes are:
\begin{itemize}
    \item \textbf{IQL}~\cite{kostrikov2022iql} updates $Q_\phi$ using the TD loss (Eq.~\ref{eq:value_loss}) and learns $V_\psi$ by $\tau$-expectile regression on the in-dataset action values $Q(s_t,\mathbf{a}_t)$. We set $\tau=0.8$ in this study.
    \item \textbf{SARSA} denotes the $\tau=0.5$ endpoint of IQL, corresponding to the SARSA-style TD backup discussed in the original IQL paper~\cite{kostrikov2022iql}. 
    The resulting $V_\psi$ regresses toward the mean $Q$-value over in-dataset actions.
    \item \textbf{ALOE}~\cite{yang2026aloe} uses a policy-sampled TD target by replacing $V_{\psi^-}(s_{t+1})$ in Eq.~\ref{eq:value_loss} with $Q_{\phi^-}(s_{t+1},\hat{\mathbf{a}}_{t+1})$, where $\hat{\mathbf{a}}_{t+1}\sim\pi(\cdot\mid s_{t+1})$. 
    The corresponding state-value estimate is obtained from policy-sampled $Q$-values rather than a separate value network.
    \item \textbf{Monte Carlo (MC)} fits $V_\psi$ directly to empirical Monte Carlo return $G_t^{\mathrm{MC}}$, a target choice also used by RECAP~\cite{pi06-star}, and does not learn an action-value function.
\end{itemize}

IQL and SARSA learn both $Q$ and $V$, whereas ALOE learns $Q$ and derives a state-value estimate from policy-sampled $Q$-values. 
We evaluate all three with both $Q-V$ scoring (Eq.~\ref{eq:advantage_qv}) and $n$-step TD scoring (Eq.~\ref{eq:advantage_n_step_td}). 
Since MC learns only $V$, we evaluate it only with $n$-step TD scoring, yielding seven configurations in total.

% The residual advances by a fixed row offset spanning an integer number of chunks and discounts its bootstrap per chunk, so we report it as a logged-continuation residual at the horizon stated in Sec.~\ref{sec:evaluation} rather than as a one-step backup.

% The TD($n$) score uses an $n$-chunk horizon and applies the discount once per action chunk.
% The corresponding replay-row offset is specified in Sec.~\ref{sec:evaluation}.
% No actor parameter changes during Stage I, and diagnostics are computed on the training replay rather than held-out data.

The estimates are intended to assign higher advantages to actions that contribute more to task completion. 
Because the compared constructions estimate different quantities, we evaluate the orderings they induce rather than numerical values. 
Episode-level outcomes are too coarse for this purpose: failed trajectories may contain actions that complete earlier sub-tasks, while successful trajectories may contain suboptimal but recoverable actions. 
We therefore use two pairwise ordering diagnostics: $R_{\mathrm{int}}$ draws on direct but event-limited evidence from human \textit{intervention}, whereas $R_{\mathrm{prog}}$ draws on broader but more indirect evidence from automated \textit{progress} segmentation and visual matching. 
We consider the two diagnostics jointly and treat their agreement as stronger screening evidence than either diagnostic alone.

The intervention diagnostic $R_{\mathrm{int}}$ pairs each human-takeover window with the equal-length policy-controlled window immediately preceding it. 
Because human operators take over when the policy is judged likely to fail, the subsequent takeover behavior is expected to receive higher advantages. 
We average the estimated advantages within each window and define $R_{\mathrm{int}}$ as the fraction of takeover events in which the takeover window receives the higher advantage.

\begin{figure*}[t]
    \vspace{5pt}
    \centering
    \includegraphics[width=1.0\linewidth]{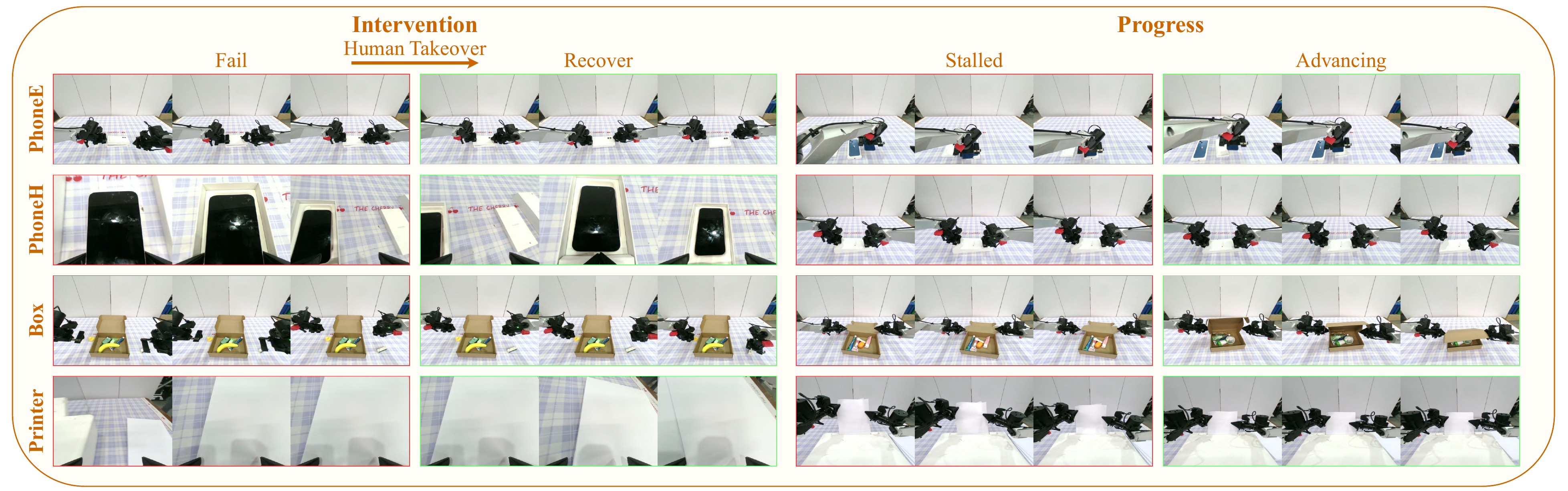}
    \caption{Representative intervention and progress pairs across four real-world tasks.
    Each strip contains three frames in temporal order.
    The intervention examples (left) show failed policy behavior (\textcolor[RGB]{255,51,51}{red}) followed by recovery after human takeover (\textcolor[RGB]{102,255,102}{green}), whereas the progress examples (right) contrast stalled behavior (\textcolor[RGB]{255,51,51}{red}) with advancing behavior (\textcolor[RGB]{102,255,102}{green}) within the same sub-task.
    Both diagnostics assess whether the green examples receive higher estimated advantages than their red counterparts.
    Images use the base-camera view except for the PhoneH and Printer intervention examples, which use the left- and right-wrist views, respectively.}
    \label{fig:sample_contrast}
    \vspace{-15pt}
\end{figure*}

The progress diagnostic $R_{\mathrm{prog}}$ compares \emph{advancing} and \emph{stalled} segments within the same sub-task. 
For each task, we manually define a sequence of $4$--$6$ sub-tasks and provide a textual description for each. The detailed sub-tasks for each task can be found in the supplementary video due to page limitations.
GPT-5.6 then segments each trajectory using these descriptions, visual observations, and robot states. All segments in successful trajectories are labeled \emph{advancing}; in failed trajectories, completed segments are labeled \emph{advancing}, while the attempted but incomplete segment is labeled \emph{stalled}. We average the estimated advantages over six one-second windows from each segment and match each stalled segment to an advancing segment using the distance between their three-camera embeddings from SmolVLM2-500M~\cite{marafioti2025smolvlm}, excluding same-trajectory matches and pairs beyond the task-level $95$th-percentile distance. $R_{\mathrm{prog}}$ is the fraction of matched pairs in which the advancing segment receives the higher advantage, averaged first within each sub-task and then across sub-tasks. For both diagnostics, $0.5$ indicates chance ordering. Fig.~\ref{fig:sample_contrast} shows representative pairs.

\subsection{Stage II: Advantage Calibration}
Stage II fixes the advantage construction retained from Stage I and varies only the calibration grouping rule $\mathcal{G}$ in Eq.~\ref{eq:score_calibration}. In long-horizon tasks with sparse terminal rewards, sub-tasks differ in state distribution and remaining horizon, so raw advantage levels may vary systematically with task progress. Calibration should reduce these context-dependent differences so that calibrated advantages reflect an action's relative quality within its decision context rather than the absolute progress of that context. Actions of comparable relative quality should therefore receive comparable advantages across sub-tasks. We assess this property by measuring systematic differences in calibrated advantage levels across sub-tasks. The sub-task partition established in Stage I is reused only for this offline diagnostic, and sub-task labels are not provided to any calibration operator. Since policy rollouts and expert demonstrations may come from different behavior distributions, normalization statistics are computed separately by source. The comparison includes an uncalibrated baseline together with global and two group-wise normalizations, a distinction also studied in LLM reinforcement learning~\cite{zhu2025stratifiedgrpo}:

\begin{itemize}
    \item \textbf{Raw} leaves the raw advantages unchanged.
    \item \textbf{Global} normalizes all samples from each source using source-level statistics as Eq.~\ref{eq:score_calibration}.
    \item \textbf{Temporal} further partitions each source into $B$ equal-width bins by episode-relative temporal position and normalizes advantages within each bin, motivated by the step-level relative advantages of TGRPO~\cite{chen2025tgrpo}.
    \item \textbf{Value} clips the estimated state values $V_{\psi^-}$ to $[0,1]$, partitions each source into $B$ equal-width bins over this value, and normalizes advantages within each bin.
\end{itemize}
For Temporal and Value, we use $B=8$ bins per source and compute all normalization statistics once from the dataset, holding them constant during Stage III. Bins with fewer than $32$ samples use the corresponding source-level Global statistics. For each source, $\sigma_{\min}$ in Eq.~\ref{eq:score_calibration} is set to $0.1$ times its global standard deviation.

We quantify these cross-sub-task differences using $\eta^2$, the ratio of variation between sub-task means to total variation in calibrated advantages:
\begin{equation}
\eta^2 = \frac{\sum_k n_k(\bar{z}_k-\bar{z})^2}{\sum_i(z_i-\bar{z})^2},
\end{equation}
where $z_i$ is the calibrated advantage of data sample $i$, with $z_i=u_i$ under Raw. $\bar{z}_k$ is the mean over the $n_k$ samples assigned to sub-task $k$, and $\bar{z}$ is the mean over all samples. A lower $\eta^2$ indicates smaller differences between sub-task means relative to the overall variation and therefore more comparable advantage levels across sub-tasks.

\subsection{Stage III: Advantage Utilization}
Stage III fixes the advantage construction and calibration retained from Stages I and II, and varies only whether and how the calibrated advantages are used in the actor update.
We compare three post-training conditions, along with their shared SFT initialization as a reference. All three conditions optimize the same $\pi_{0.5}$ flow-matching actor~\cite{black2025pi05} through Eq.~\ref{eq:weighted_actor_loss} using the same dataset and training budget, and differ only in the weight $w_i$ assigned to a data sample.

\begin{itemize}
    \item \textbf{SFT Init} is the shared initialization, trained on the demonstrations alone and evaluated without any further actor update.
    \item \textbf{DAgger} uses an advantage-free binary rule following HG-DAgger~\cite{ross2011dagger,kelly2019hgdagger}: human samples, including both demonstrations and takeover segments, receive unit weight, while autonomous rollouts receive zero weight.
    \item \textbf{Weight} follows the principle of AWR~\cite{peng2020awr} and assigns each sample a continuous exponential weight $w_i=\exp\left(\operatorname{clip}(\beta z_i,-c,c)\right)$, where $z_i$ is the calibrated advantage, with $\beta=1$ and $c=2$ in our experiments.
    \item \textbf{Filter} assigns unit weight to the samples whose calibrated advantages are above the $q$-quantile of the training batch and zero weight to the remaining samples, with $q=0.5$.
    This follows the hard-filtering rule of Imit-RECAP baseline in HABC~\cite{fang2026habc}.
\end{itemize}

We do not include RECAP itself as a condition: its advantage conditioning is introduced during pre-training~\cite{pi06-star}, so a post-training-only re-implementation departs from the original recipe and has been reported to yield more modest gains over supervised fine-tuning~\cite{wang2026lwd}. Filter instead serves as a hard approximation that discards low-advantage samples rather than conditioning on them.

Each comparison supports a different attribution.
Weight against Filter isolates the weighting form, because both consume the same advantages and differ only in whether the weight is continuous or binary.
Weight and Filter against DAgger compare weighting by estimated action quality with weighting by data source, and all three against SFT Init test whether post-training improves over the initialization.

% \subsection{Policy Evaluation}

% At deployment, all policies receive the same observations and use the same action-chunk interface, and no critic is queried at inference time.
% For each real-world task, we report the mean task progress and the success rate over all evaluation trials.
% A trial terminates on successful completion, on an unrecoverable failure, or at a timeout, where the timeout is the longest expert-episode duration for the corresponding task.
% We treat as unrecoverable failures moving an object outside the reachable workspace and starting a later sub-task before its prerequisite is complete.
% Task progress is the fraction of the sub-tasks of that task, defined as in Stage I, that an operator judges to be complete when the trial terminates.
% Task success is $1$ if and only if task progress reaches $1$, so it is the strictest point of the same ordered scale.

\section{Experiment Results}
\label{sec:evaluation}

\begin{figure*}[t]
\vspace{5pt}
\centering
\includegraphics[width=0.98\linewidth]{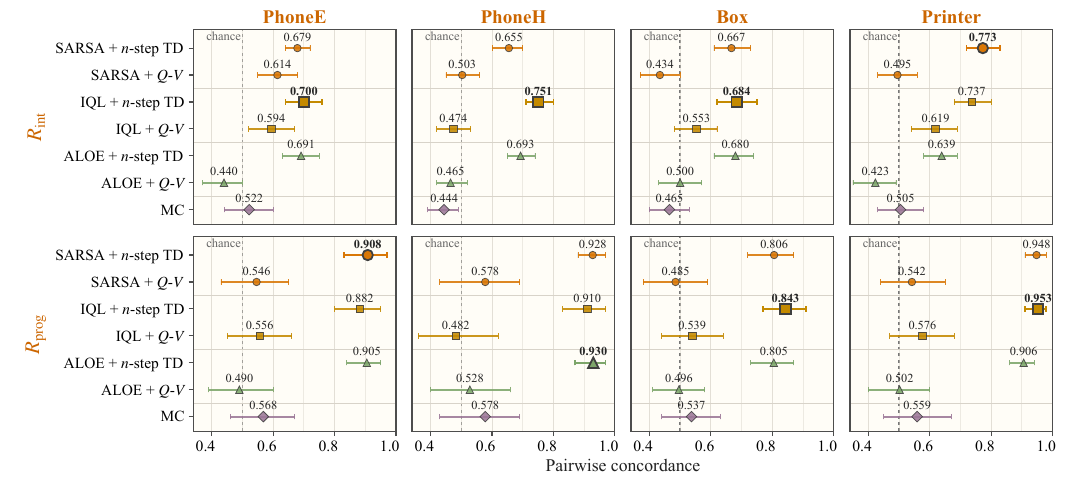}
\caption{
Advantage-construction diagnostics computed across four real-world tasks. 
The top row reports intervention ordering $R_{\mathrm{int}}$ and the bottom row reports progress ordering $R_{\mathrm{prog}}$.
For each task, the intervention and progress diagnostics use approximately $200$ and $400$ paired comparisons, respectively, after applying their diagnostic-specific filtering rules.
Point markers show the ordering diagnostics (higher is better), and horizontal bars show $95\%$ confidence intervals. 
The dashed line marks chance ordering at $0.5$.
\textbf{Bold} numeric labels with heavier marker outlines identify the within-panel maxima.
}
\label{fig:rq1_diagnostics}
\vspace{-15pt}
\end{figure*}

Following the protocol in Sec.~\ref{sec:method}, we first screen advantage construction and calibration offline, then evaluate policy utilization on four real-world tasks.
We also revisit alternative construction and calibration choices in real-world ablations.
% and evaluate the selected recipe on \textit{LIBERO}.

\subsection{Experimental Setup}

\noindent\textbf{Real-world tasks and platform.}
% We study four bimanual manipulation tasks, of which representative frames are shown in Fig.~\ref{fig:realscene}.
We study four bimanual manipulation tasks spanning object insertion, long-horizon sequencing, multi-object interaction, and deformable-object handling (Fig.~\ref{fig:realscene}).

\begin{itemize}
    \item \textit{Pack Phone Easy (PhoneE)} places a phone into its packaging box and closes its lid.
    \item \textit{Pack Phone Hard (PhoneH)} uses a more complex packaging box and requires an accessory-insertion step, resulting in a longer-horizon manipulation task with greater overall difficulty.
    \item \textit{Pack Box (Box)} places three objects into a carton, folds the flaps, and presses the side tabs to close the carton.
    \item \textit{Refill Printer Paper (Printer)} picks up a small stack of paper and loads it into the printer tray.
\end{itemize}
% Note that the two phone-task names are operational identifiers for distinct packaging instances, not a claim that the tasks differ only by difficulty or share an object model.    
%\noindent These tasks cover dexterous manipulation (PhoneE\&H), multi-object interaction (Box), and deformable-object handling (Printer) to evaluate the policies comprehensively.
The bimanual robot platform comprises two 6-DoF arms, each equipped with a gripper, and three RGB fisheye cameras: one mounted on the robot base and one on each wrist.
At each policy query, the policy receives the end-effector states and three undistorted camera images, and predicts a chunk of $C=30$ end-effector actions.
The robot executes the chunked actions at $30$~Hz, completing the chunk in one second before querying the locally deployed policy again.
All compared policies are evaluated using the same robot embodiment, observation and action spaces, deployment interface, and evaluation initial-condition schedule.
% TODO: Freeze the operational real-world task definitions, camera specifications, hardware, latency, and LIBERO tasks and episodes.

\noindent\textbf{Dataset and implementation details.}
For each task, we construct a fixed offline dataset consisting of approximately $40$ hours of expert demonstrations used to train the task-specific SFT initialization, $100$ autonomous rollout episodes, and $100$ rollout episodes with human takeovers.
All critic-learning and advantage-construction experiments are conducted on this fixed dataset. We implement all training procedures based on the LeRobot codebase~\cite{cadene2024lerobot}. Specifically, each learned $Q$ or $V$ function is parameterized by a separate encoder-only Transformer followed by an MLP output head, and reuses the visual embedding of the VLA policy~\cite{yang2026aloe}.
For $n$-step TD advantage estimation, we use a per-chunk discount factor of $\gamma=0.96$ and a backup horizon of $n=10$ chunks.
Because each chunk contains $C=30$ primitive actions and spans one second, the backup covers $nC=300$ primitive actions, or 10 seconds of execution.
% Our RFT implementation is mainly based on LeRobot~\cite{cadene2024lerobot}. 
% The primary policy comparisons use at least three paired pipeline seeds and the matched initialization and training settings described in Sec.~\ref{sec:method}.
% TODO: Freeze replay source proportions, critic and actor updates, global batch size, seeds, and checkpoint-selection rule.

% \paragraph{Analysis protocol.}
% The pipeline seed is the experimental unit for training variation, while tasks and evaluation episodes provide lower-level observations.
% We report paired effects with 95\% intervals, real-world setting and task-family breakdowns, LIBERO suite breakdowns, and absolute success-rate differences in percentage points.
% RQ1 and RQ2 are descriptive offline studies, whereas RQ3 uses ordered milestone progress and binary success as co-reported real-world endpoints.
% Within RQ3, C2 versus C3 is the matched utilization contrast under the selected score--calibration configuration, and C2 versus S1 or S2 measures sensitivity to the score--calibration choice under AWR.
% Real-world success is analyzed within task, day, checkpoint, and initialization blocks; trial counts and the blocked analysis schedule are fixed before deployment.
% Intervals spanning both negligible and meaningful effects are reported as inconclusive rather than as evidence of equivalence.
% TODO: Freeze endpoint-specific practical-equivalence margins, the interval estimator, multiplicity families, real-world trial counts, and the blocked analysis model before actor outcomes are inspected.
\subsection{RQ1: Advantage Construction}

\begin{table}[t]
    \centering
    \caption{
    Stage-II offline calibration results measured by $\eta^2$, the proportion of total advantage variance explained by sub-task identity.
Lower $\eta^2$ indicates less sub-task-dependent variation; Mean is the average across the four tasks.
    % and \emph{LevelSpread} denotes the difference between the
    % highest and lowest sub-task medians normalized by the global standard deviation.
    % The final column reports the equally weighted mean across the four tasks.
    }
    \label{tab:calibration_scale}
    \small
    \setlength{\tabcolsep}{4pt}
    \begin{tabular}{lccccc}
        \toprule
        Calibration
        & PhoneE
        & PhoneH
        & Box
        & Printer
        & Mean \\
        \midrule
        % \multicolumn{6}{l}{\textit{Between-sub-task variance share} $\eta^2$} \\
        \textbf{Raw}
        & 0.072 & 0.063 & 0.106 & 0.127 & 0.092 \\
        \textbf{Global}
        & 0.057 & 0.051 & 0.090 & 0.120 & 0.080 \\
        \textbf{Temporal}
        & 0.071 & \textbf{0.023} & 0.053 & 0.028 & 0.044 \\
        \textbf{Value}
        & \textbf{0.053} & 0.027 & \textbf{0.013} & \textbf{0.010} & \textbf{0.026} \\
        % \midrule
        % \multicolumn{6}{l}{\textit{LevelSpread}} \\
        % \textbf{Raw}
        % & 0.448 & 0.685 & 0.425 & 2.327 & 0.971 \\
        % \textbf{Global}
        % & 0.448 & 0.685 & 0.425 & 2.327 & 0.971 \\
        % \textbf{Temporal}
        % & 0.599 & 0.917 & 0.338 & 1.494 & 0.837 \\
        % \textbf{Value}
        % & 0.504 & 1.184 & 0.433 & 0.896 & 0.754 \\
        \bottomrule
    \end{tabular}
    \vspace{-10pt}
\end{table}

\begin{table*}[tb]
\centering
\vspace{5pt}
\caption{
Real-world policy evaluation for RQ3.
Values in the task columns are averages over $20$ real-robot trials per condition; Mean is the average across the four tasks. Higher is better; bold denotes the highest in each column.}

\label{tab:real_policy_results}
\small
\setlength{\tabcolsep}{4pt}
\renewcommand{\arraystretch}{1.0}
    \begin{tabular*}{0.8\textwidth}{@{\extracolsep{\fill}}l*{10}{c}@{}}
        \toprule
        Condition
        & \multicolumn{5}{c}{Task progress}
        & \multicolumn{5}{c}{Task success} \\
        \cmidrule(lr){2-6} \cmidrule(lr){7-11}
        & PhoneE & PhoneH & Box & Printer & Mean
        & PhoneE & PhoneH & Box & Printer & Mean \\
        \midrule
        \textbf{SFT Init}
        & 0.50 & 0.32 & 0.69 & 0.24 & 0.44
        & 0.30 & 0.00 & 0.15 & 0.00 & 0.11 \\
        \textbf{DAgger}
        & 0.65 & 0.62 & 0.91 & 0.66 & 0.71
        & 0.35 & 0.30 & 0.70 & 0.45 & 0.45 \\
        \textbf{Weight}
        & \textbf{0.88} & \textbf{0.81} & \textbf{0.95}
        & \textbf{0.79} & \textbf{0.86}
        & \textbf{0.80} & \textbf{0.70} & \textbf{0.80}
        & \textbf{0.65} & \textbf{0.74} \\
        \textbf{Filter}
        & 0.78 & 0.69 & 0.91 & 0.60 & 0.75
        & 0.70 & 0.40 & 0.55 & 0.35 & 0.50 \\
        \bottomrule
    \end{tabular*}
    \vspace{-15pt}
\end{table*}

We evaluate the seven advantage-construction configurations using the intervention diagnostic $R_{\mathrm{int}}$ and progress diagnostic $R_{\mathrm{prog}}$, defined in Sec.~\ref{sec:method} and illustrated in Fig.~\ref{fig:sample_contrast}.
All critic checkpoints are evaluated after $10{,}000$ training updates with batch size $2048$.
The results are shown in Fig.~\ref{fig:rq1_diagnostics}.
For IQL, SARSA, and ALOE critics, $n$-step TD yields higher concordance than $Q-V$ on every task and both diagnostics.
% We evaluate advantage construction using two pairwise ordering diagnostics, intervention ordering $R_{\mathrm{int}}$ and progress ordering $R_{\mathrm{prog}}$.
% The former compares human-takeover windows with the immediately preceding, equal-length policy windows, while the latter compares visually matched advancing and stalled units from the same sub-task.
% Both measure agreement with the expected advantage ordering, with $0.5$ indicating chance ordering.
% Using these diagnostics, we compare seven advantage-estimation configurations across the four tasks, with results shown in Fig.~\ref{fig:rq1_diagnostics}.% and Table~\ref{tab:score_diagnostics}.
% The $n$-step TD variants of IQL, SARSA, and ALOE-style exceed chance on both diagnostics for every task.
These three $n$-step TD variants exceed chance throughout, with $R_{\mathrm{int}}$ ranging from $0.639$ to $0.773$ and $R_{\mathrm{prog}}$ from $0.805$ to $0.953$.
In contrast, the $Q-V$ variants and the MC baseline are generally closer to chance.

We hypothesize that the weak $Q-V$ results stem from the limited action diversity, which is common in real-robot datasets collected from a small set of behavior sources.
Under narrow action coverage, $V$ is derived from $Q$ evaluated at the same or a similar action, so their subtraction cancels much of the shared value signal and leaves a small residual susceptible to estimation noise.
This issue also motivates prior work to deliberately collect diverse failure actions for critic learning~\cite{wang2026lwd}.
In contrast, $n$-step TD uses rewards and value changes along logged continuations without relying on action discrimination by $Q$.
MC remains weaker since fitting $V$ to realized returns may leave only small TD residuals along the same logged continuations.

% Because MC also uses $n$-step TD, its weaker performance indicates that the value-learning scheme matters alongside the advantage estimate.
% Their $R_{\mathrm{int}}$ values range from $0.639$ to $0.773$, and their
% $R_{\mathrm{prog}}$ values range from $0.805$ to $0.953$.
% In contrast, the $Q-V$ and MC configurations are generally weaker and
% closer to chance, although some task--configuration pairs remain above $0.5$.
% Overall, the results support $n$-step TD as the more reliable advantage
% construction, without showing a single critic that dominates across all tasks
% and diagnostics. We therefore use IQL with $n$-step TD in the subsequent stages.

Among the $n$-step TD variants, IQL achieves the highest four-task mean $R_{\mathrm{int}}$ ($0.718$), while its mean $R_{\mathrm{prog}}$ ($0.897$) nearly matches the best value achieved by SARSA ($0.898$).
We therefore retain IQL with $n$-step TD, whose upper-expectile regression can emphasize high-value in-dataset actions, a desirable property for heterogeneous datasets~\cite{kostrikov2022iql}.

\subsection{RQ2: Advantage Calibration}

Stage II evaluates whether calibration reduces systematic differences in advantage levels across sub-tasks, using IQL with the $n$-step TD construction retained from Stage I.
Calibration statistics are computed separately within each data source, following the source-wise grouping rules in Sec.~\ref{sec:method}.
The $\eta^2$ diagnostic is then computed accordingly within each task.
It measures the proportion of total advantage variance explained by sub-task identity, with lower values indicating greater comparability across sub-tasks.

As shown in Table~\ref{tab:calibration_scale}, 
all three calibration methods reduce the mean $\eta^2$ from $0.092$ for Raw to $0.080$, $0.044$, and $0.026$ for Global, Temporal, and Value, respectively. 
Compared with Global normalization, the two group-wise normalization variants further reduce the mean $\eta^2$ by removing group-specific location and scale differences before the sub-task means are compared.
Value achieves a lower mean $\eta^2$ than Temporal. One possible explanation is that variations in execution speed, stalls, and recoveries cause temporal bins to mix different progress levels, whereas state value more closely tracks task progress.
We therefore carry Value calibration into Stage III and revisit Global and Temporal calibration in the real-world ablations.

% because stalled and recovery segments can place multiple sub-tasks in the same temporal bin, whereas the learned state value is more closely aligned with task progress in these trajectories and therefore yields bins that better follow sub-task boundaries.
% We therefore carry Value calibration into Stage III and revisit Global and Temporal calibration in the real-world ablations.
% These results provide offline screening evidence about advantage comparability rather than direct evidence of policy improvement.

\subsection{RQ3: Advantage Utilization}
We next evaluate three post-training conditions together with the shared SFT initialization as a reference, using the advantage configuration selected in RQ1 and RQ2. Each post-training condition further trains the SFT checkpoint with $4000$ updates and batch size $2048$. We report task progress, the fraction of completed sub-tasks at termination, and task success, attained only when all sub-tasks are completed.
A trial terminates upon success, an unrecoverable failure, or a task-specific timeout equal to the duration of the longest expert episode.
We consider a failure unrecoverable if an object becomes unreachable or the robot begins a later sub-task before completing an earlier one.

% Using the IQL $n$-step TD construction and Value calibration selected in RQ1 and RQ2, we evaluate four policy-training conditions on each real-world task.
% At deployment, all policies receive the same observations and use the same action-chunk interface, and no critic is queried.
% For each task, we report the mean task progress and the success rate over all evaluation trials.
% A trial terminates upon successful completion, an unrecoverable failure, or a timeout set to the longest expert-episode duration for that task.
% We treat moving an object outside the reachable workspace and starting a later sub-task before its prerequisite is complete as unrecoverable failures.
% Task progress is the fraction of sub-tasks judged complete by an operator when the trial terminates, and task success is one if and only if task progress reaches one.

\noindent\textbf{Overall performance.} 
Table~\ref{tab:real_policy_results} reports the real-world task performance. All three post-training conditions yield higher four-task means for both progress and success than SFT Init. 
% Among them, Weight achieves the highest point estimates for both metrics, with means of $0.86$ for progress and $0.74$ for success, corresponding to gains of $0.42$ and $0.63$ over SFT Init, respectively.
Among them, Weight achieves the highest point estimates, gaining $0.42$ in progress and $0.63$ in success over SFT Init.
Weight's finer-grained sample-level supervision may contribute to these gains: it emphasizes higher-scoring actions while retaining nonzero weights for the remaining samples, reducing sensitivity to occasional critic misranking. Qualitatively, Weight exhibits more consistent recovery behaviors in real-robot trials, as shown in the supplementary video.

% Table~\ref{tab:real_policy_results} reports the real-world task performance.
% Weight achieves the highest point estimate for both metrics on all four tasks, with a mean of $0.86$ for task progress and $0.74$ for task success, outperforming DAgger and Filter on all tasks.
% All three post-training conditions using dataset improve both mean metrics over SFT Init,
% with Weight yielding the largest gains of $0.42$ and $0.63$, respectively.
% These gains suggest that continuously emphasizing estimated higher-quality actions within each source, while retaining the remaining supervision, exploits heterogeneous dataset more effectively than source-based DAgger or binary Filter.

\begin{figure}[tb]
    \centering
    \includegraphics[width=0.99\linewidth]{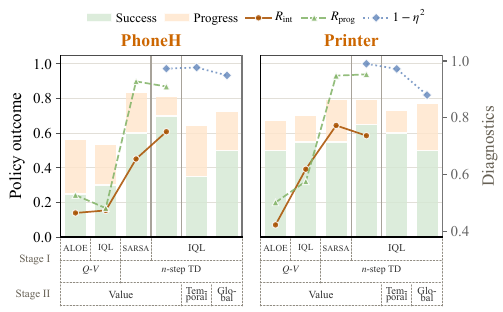}
    \caption{
    Ablations of advantage construction and calibration under fixed Weight utilization, with corresponding offline diagnostics overlaid (higher is better).
    The first four conditions vary Stage-I construction and the last three vary Stage-II calibration; the shared fourth condition is the reference configuration reported in Table~\ref{tab:real_policy_results}.
    }
    \label{fig:ablation}
    \vspace{-18pt}
\end{figure}

\noindent\textbf{Ablation study}.
% Keeping Weight fixed, we vary the Stage-I construction and Stage-II calibration.
% As shown in Fig.~\ref{fig:ablation}, replacing the selected $n$-step TD construction with $Q-V$ generally lowers both metrics, while replacing Value calibration with Global or Temporal also lowers performance.
% These ablations are consistent with the offline selections, suggesting that advantages that better identify useful behavior and remain comparable across sub-tasks provide stronger supervision for advantage-guided learning.
Keeping Weight fixed, we vary the Stage-I construction and Stage-II calibration.
As shown in Fig.~\ref{fig:ablation}, replacing the selected $n$-step TD construction with $Q-V$ generally lowers both metrics, while replacing Value calibration with Global or Temporal also degrades performance in most cases.
Across the tested ablations, higher Stage-I ordering concordance and lower Stage-II $\eta^2$ generally correspond to higher real-world task progress and success.
% except for the comparison between Value and Temporal on PhoneH.
This overall alignment supports using the diagnostics to screen design choices offline, reducing the need to evaluate every configuration on real robots.

\section{Conclusion}
\label{sec:conclusion}
% We presented a staged empirical study of advantage-guided VLA post-training with limited and heterogeneous robot replay, separating advantage construction, calibration, and utilization.
% Stage-specific offline diagnostics selected IQL with chunk-level $n$-step TD advantages and value-based calibration, while real-robot evaluation favored continuous advantage weighting over source-based DAgger and hard filtering.
% Across four bimanual tasks, the resulting recipe produced the highest point estimates for both task progress and success on every task, with four-task means of $0.86$ and $0.74$---gains of $0.42$ and $0.63$ over SFT Init.
% Construction and calibration ablations generally aligned with the offline diagnostics, supporting their use to narrow the post-training design space before costly real-robot evaluation.

We dissected advantage-guided VLA post-training into construction, calibration, and utilization through a staged empirical study with limited and heterogeneous robot datasets.
Stage-specific offline diagnostics selected IQL with chunk-level $n$-step TD advantages and value-based calibration, while real-robot evaluation favored continuous advantage weighting over source-based DAgger and hard filtering.
Across four bimanual tasks, the resulting recipe produced the highest point estimates for both task progress and success on every task, with four-task means of $0.86$ and $0.74$---gains of $0.42$ and $0.63$ over SFT Init, respectively.
Construction and calibration ablations generally aligned with the offline diagnostics, supporting their use to narrow the post-training design space before costly real-robot evaluation.

\bibliographystyle{IEEEtran}
\bibliography{reference}
% \printbibliography
\end{document}